\documentclass{article}

\usepackage{microtype}
\usepackage{graphicx}
\usepackage{subfigure}
\usepackage{booktabs} \usepackage{xspace}
\usepackage{enumitem}
\usepackage{fontsize}

\usepackage{hyperref}

\usepackage[accepted]{icml2023}

\usepackage{amsmath}
\usepackage{amssymb}
\usepackage{mathtools}
\usepackage{amsthm}
\usepackage{listings}

\usepackage[capitalize,noabbrev]{cleveref}

\theoremstyle{plain}

\theoremstyle{definition}

\theoremstyle{remark}

\usepackage[textsize=tiny]{todonotes}
\usepackage{verbatim}

\newcommand{\dream}{\textsc{Dream}\xspace}
\newcommand{\import}{\textsc{Import}\xspace}
\newcommand{\rl}{RL$^2$}
\newcommand{\task}{{\mu}}

\icmltitlerunning{Simple Embodied Language Learning as a Byproduct of Meta-RL}

\begin{document}

\twocolumn[
\icmltitle{Simple Embodied Language Learning as a\\ Byproduct of Meta-Reinforcement Learning}

\icmlsetsymbol{equal}{*}

\begin{icmlauthorlist}
\icmlauthor{Evan Zheran Liu}{stanford}
\icmlauthor{Sahaana Suri}{stanford}
\icmlauthor{Tong Mu}{stanford}
\icmlauthor{Allan Zhou}{stanford}
\icmlauthor{Chelsea Finn}{stanford}
\end{icmlauthorlist}

\icmlaffiliation{stanford}{Department of Computer Science, Stanford University, Stanford, CA}

\icmlcorrespondingauthor{Evan Z. Liu}{evanliu@cs.stanford.edu}

\icmlkeywords{Machine Learning, ICML}

\vskip 0.3in
]

\printAffiliationsAndNotice{\icmlEqualContribution}

\begin{abstract}
Whereas machine learning models typically learn language by directly training on language tasks (e.g., next-word prediction), language emerges in human children as a byproduct of solving non-language tasks (e.g., acquiring food).
Motivated by this observation, we ask: can embodied reinforcement learning (RL) agents also indirectly learn language from non-language tasks?
Learning to associate language with its meaning requires a dynamic environment with varied language.
Therefore, we investigate this question in a multi-task environment with language that varies across the different tasks.
Specifically, we design an office navigation environment, where the agent's goal is to find a particular office, and office locations differ in different buildings (i.e., tasks).
Each building includes a floor plan with a simple language description of the goal office's location, which can be visually read as an RGB image when visited.
We find RL agents indeed are able to indirectly learn language.
Agents trained with current meta-RL algorithms successfully generalize to reading floor plans with held-out layouts and language phrases, and quickly navigate to the correct office, despite receiving no direct language supervision.
\end{abstract}

\section{Introduction}\label{sec:intro}

\begin{figure}[t]
\begin{center}
\centerline{\includegraphics[width=\columnwidth]{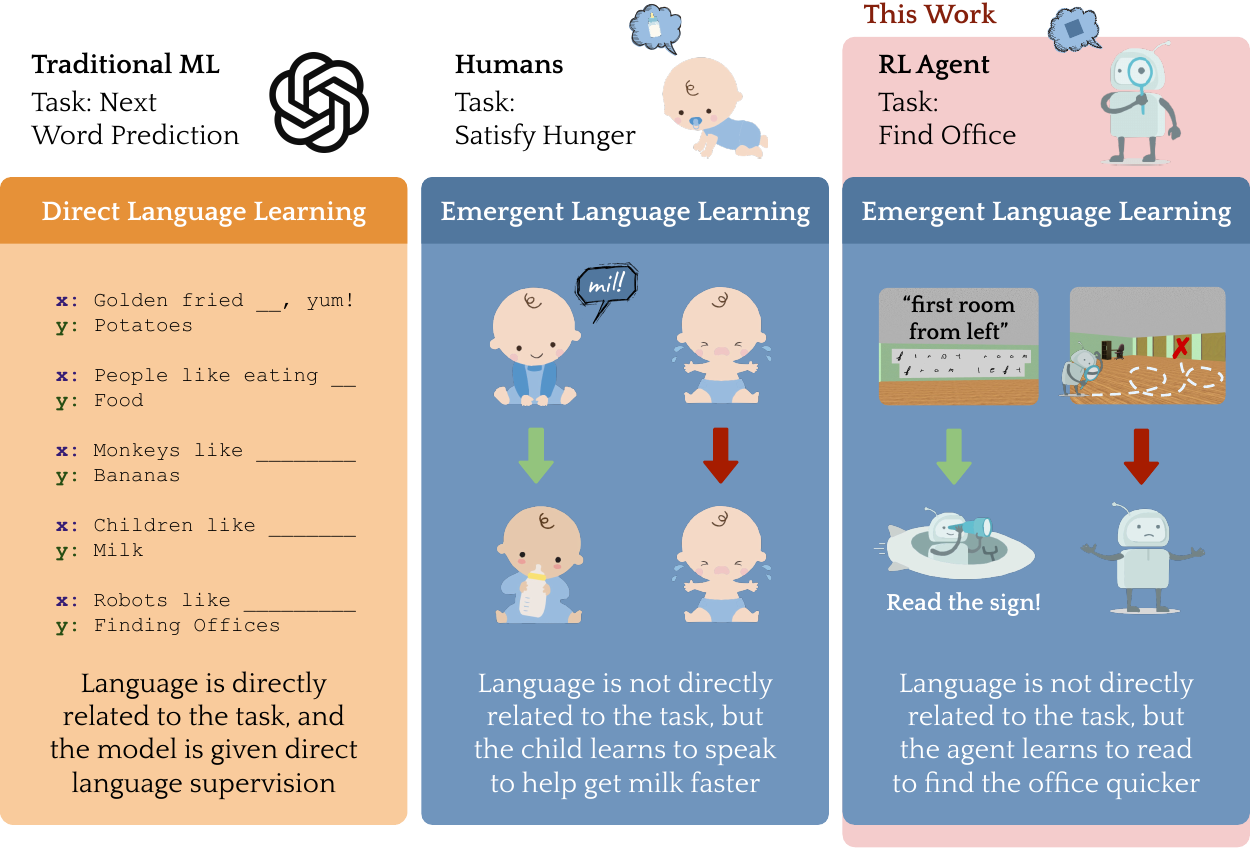}}
\vspace{-2mm}
\caption{Machine learning models
learn language by directly training on language tasks, whereas children learn language as a byproduct of solving tasks that do not explicitly relate to language.
We ask if embodied reinforcement learning agents can learn language in a similarly indirect manner.
}
\vspace{-7mm}
\label{fig:1}
\end{center}
\end{figure}

The standard paradigm for learning language with machine learning (ML) models is to train directly on language tasks, such as next-word prediction~\cite{devlin2019bert,radford2018improving,brown2020gpt3}. 
In contrast, humans \emph{indirectly} learn language as a byproduct of fulfilling non-language objectives.
For instance, children incrementally learn words that alert parents to their needs (e.g., being hungry or thirsty) from vocalizations (e.g., crying) to more effective words and sentences (e.g., ``I'm hungry").

Indirectly learning language can be desirable as it is directly tied to real-world observations, which yields language understanding that is grounded and consistent with reality. 
In contrast, standard, directly-trained models can output linguistically correct, but factually incorrect sentences~\citep{lin2021truthfulqa}.
Motivated by this observation, we investigate if embodied ML agents situated in simple 3D environments can start to learn language via indirect mechanisms similar to humans~(Figure~\ref{fig:1}).

Our investigation builds on two directions of prior work.
First, numerous studies incorporate language into reinforcement learning (RL) agents, e.g., by training them to associate language with objects~\citep{hill2020grounded}, follow language instructions~\citep{misra2017mapping,hill2020human},
and learn from language feedback~\citep{ling2017teaching}.
We similarly seek agents with language skills, but differ in the learning mechanism:
whereas the task objective either provides language supervision or directly requires language learning in the aforementioned line of work,
we instead investigate whether RL agents can learn language \emph{indirectly} without explicit language supervision.
Put another way, we study if language learning can arise as a \emph{byproduct} of learning tasks that do not necessarily require language skills, in contrast to tasks that \emph{prescribe} language learning, such as language instruction following.
Second, this mechanism of indirectly learning behaviors that are not directly prescribed by the objective of the training task, sometimes called \emph{emergence},
is also studied in other works~\citep{bansal2017emergent}.
However, while these works study the emergence of tool usage~\citep{baker2019emergent} and high-level economic policies~\citep{zheng2020ai},
we instead focus on the emergence of language, specifically reading in embodied environments.

A primary challenge of our investigation is to design an environment that enables and tests the emergence of language.
We identify four criteria for such an environment:
(1) The environment must include language. (2) The language must benefit the agent in solving the task, so the agent is incentivized to learn language.
Note that this requires the environment to vary and for some of these variations to be captured in the language.
Otherwise, the language would always convey the same information,
which would not encourage language understanding.
(3) The task must be solvable without language understanding, as tasks that \emph{require} language skills prescribe the use of language, rather than enable it to emerge as a byproduct.
(4) Observations of the language should be unprocessed to more accurately model the real world, e.g., visual observations of a sign, rather than tokenized strings.
Since, to the best of our knowledge, no existing environments meet these criteria, we design our own (Section~\ref{sec:setup}).

To satisfy the above criteria, we design an office navigation environment where the goal is to visit a specific office as quickly as possible.
To address (1), the office building contains a floor plan that indicates the location of the goal office, via language or pictorially.
While the inclusion of a floor plan partially satisfies (2), with a static environment, the agent can simply memorize the location of the target office.
Thus, to address the variations entailed by (2), we randomize the goal office location, updating the floor plan accordingly, and encode each randomization as a different task in a multi-task setting.
In addition, to enable generalization to new but related tasks and language, we give the agent a few episodes to adapt to each new task, which yields the few-shot meta-RL setting~\citep{duan2016rl,wang2016learning,finn2017modelagnostic}.
To address (3), we ensure that the agent can find the target office even without the floor plan, i.e., through exhaustive search.
Finally, to address (4), we represent the floor plan as an RGB image that the agent must visually observe and learn to parse in order to read.

In this environment, we find that the \dream meta-RL algorithm~\citep{liu2021dream} learns an agent that navigates to and reads the floor plan from raw RGB observations
during the few adaptation episodes in each new building, starting from no language knowledge a priori, and while receiving no special reward bonuses for reading the floor plan.
The agent then uses the information from the floor plan to immediately navigate to the goal office.
We also find that the learned agent exhibits a degree of compositional generalization to new language phrases in the floor plan, indicating the emergence of simple language skills (Section~\ref{sec:experiments}).

Overall, our main contribution is raising and providing an affirmative answer to the question \emph{can simple language skills emerge in meta-RL agents without direct language supervision?}
Additionally, we vary the meta-RL learning algorithm and parameters of our open-source office environment to determine factors that affect whether language arises.
We find that language skills arise proportionally to the benefit they provide to the agent:
When there are only a few buildings, which require visiting only a small number of offices to disambiguate, agents are more likely to visit those offices rather than develop language skills.
In contrast, when there are many buildings, which require visiting many offices to disambiguate, agents more often develop language skills.

\section{Related Work}\label{sec:related_work}

Machine learning models for language are typically trained on tasks with explicit language supervision, in the form of text examples with labels~\citep{zhang2018deep,li2020survey,devlin2019bert,brown2020gpt3,vaswani2017attention}.
We instead ask if language can be learned \emph{indirectly} in the process of solving RL tasks that do not provide such supervision.
This question combines elements from two groups of prior work:

\textbf{Language in reinforcement learning.}
A rich literature incorporates language into RL agents situated in the world.
These works include language in ways that span text-based games~\citep{narasimhan2015language,yuan2018counting,kuttler2020nethack},
grounding tasks that require associating words with objects in the world~\citep{hermann2017grounded,hill2020grounded,yan2022intra,ahn2022saycan},
language instruction following~\citep{shah2018follownet,hill2020human,vaezipoor2021ltl2action},
language feedback~\citep{ling2017teaching,kreutzer2020offline},
inducing rewards from a language goal~\citep{fu2019language,sumers2021learning},
and leveraging language as an abstraction for hierarchical RL~\citep{andreas2017learning,jiang2019abstraction}.
For a complete survey, see~\citet{luketina2019survey}.
We similarly study language in RL, but rather than rely on explicit language supervision like these studies or tasks \emph{requiring} language skills, 
we study language learning in tasks without direct language supervision, which can be solved without any language.
In other words, language learning is a direct consequence of the task in these works (e.g., instruction following prescribes language learning), while our tasks do not necessarily require language to solve, and hence allow for language to arise as a \emph{byproduct}.

\textbf{Emergent behaviors and capabilities.}
A long line of work studies the emergence of behaviors that are not directly specified by the objective of the task.
This phenomenon has primarily been observed in the multi-agent setting~\citep{bansal2017emergent,baker2019emergent,jaderberg2019human,zheng2020ai,team2021open,team2023human}, where behaviors that are not directly rewarded by the task (e.g., using tools, developing economic specializations, and communicating) arise due to pressures from cooperation and competition.
We also focus on indirectly learning behaviors not specified by the task, but differ as prior work has not studied the emergence of language.
Notably, \citet{brown2020gpt3,wei2022emergent} also study emergent generalization to new language tasks within large language models, though these models are still trained with explicit language supervision.

\section{Preliminaries}
\label{sec:preliminaries}

\begin{figure}[t]
\begin{center}
\centerline{\includegraphics[width=0.98\columnwidth]{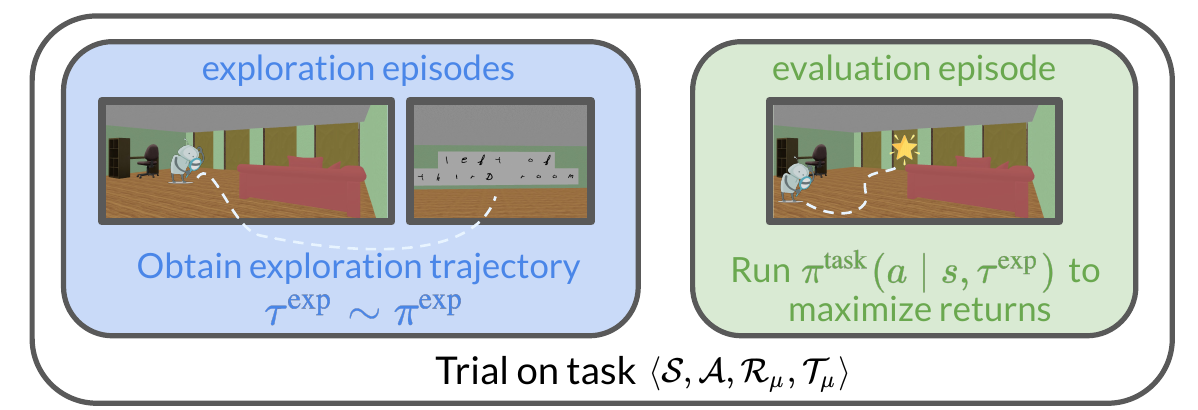}}
\caption{Illustration of a meta-RL trial.
A trial consists of two parts:
(i) running an exploration policy $\pi^\text{exp}$ on the exploration episodes (i.e., the few shots);
(ii) running a task policy $\pi^\text{task}$ on the evaluation episode, conditioned on what was discovered during the exploration episodes.
The agent is evaluated based on the returns achieved during the evaluation episode, while the returns achieved during the exploration episodes do not matter.}
\vspace{-5mm}
\label{fig:setting}
\end{center}
\end{figure}

\subsection{Meta-Reinforcement Learning Setting}\label{sec:meta_rl_setting}
To create a dynamic environment with varied language, we consider the few-shot meta-RL setting.
At a high level, the goal in this setting is to learn an agent that can solve new tasks in only a few shots (i.e., episodes), after training on related tasks.

More formally, the meta-RL setting consists of a distribution over tasks $p(\task)$, where each task $\task$ is a Markov decision process $\langle \mathcal{S}, \mathcal{A}, \mathcal{R}_\task, T_\task\rangle$ with states $\mathcal{S}$, actions $\mathcal{A}$, reward function $\mathcal{R}_\task$ and dynamics $T_\task$.
We formalize learning from a few shots on each task as a \emph{trial}~\citep{duan2016rl,liu2021dream}, as in Figure~\ref{fig:setting}.
A trial consists of sampling a new task $\task \sim p(\task)$, running a policy $\pi^\text{exp}$ for a few \emph{exploration episodes}, which yields a trajectory $\tau^\text{exp} = (s_0, a_0, r_0, \ldots)$, and running a policy $\pi^\text{task}$ on a final \emph{evaluation episode} conditioned on the exploration trajectory $\tau^\text{exp}$.
Note that $\pi^\text{exp}$ and $\pi^\text{task}$ can be the same policy.
The goal of the meta-RL setting is to learn policies $\pi^\text{exp}$ and $\pi^\text{task}$ that maximize the expected returns achieved by $\pi^\text{task}$ over trials:
\begin{align}\label{eqn:meta_rl_objective}
    \mathbb{E}_{\task \sim p(\task), \tau^\text{exp} \sim \pi^\text{exp}(\task)} \left[V_\pi^\text{task} \left(\tau^\text{exp} \right)\right],
\end{align}
where $V^\text{task}_\pi(\tau^\text{exp})$ denotes the expected returns of $\pi^\text{task}$ conditioned on $\tau^\text{exp}$ on a single episode, and $\pi^\text{exp}(\task)$ represents the distribution over trajectories from running $\pi^\text{exp}$ on $\mu$.

Additionally, several algorithms we use assume that the tasks sampled during meta-training (but not meta-testing) are identifiable (i.e., each meta-training task is assigned a unique one-hot ID).
We slightly overload notation and also let $\task$ denote this unique one-hot ID.

\subsection{Meta-Reinforcement Algorithms}\label{sec:meta_rl_setting}
\textbf{End-to-end approaches.} A canonical approach for the meta-RL setting is to learn a single policy $\pi^\text{exp} = \pi^\text{task} = \pi$ trained end-to-end to directly maximize the meta-RL objective (Equation~\ref{eqn:meta_rl_objective}).
This can be achieved by grouping the exploration episodes and evaluation episodes together as a single long episode and training with standard RL, where the policy $\pi$ is made recurrent to condition on $\tau^\text{exp}$ during the evaluation episode portion.
The most basic version of this approach is known as \rl~\citep{duan2016rl,wang2016learning}, though several approaches propose adding auxiliary objectives to improve learning.

For example, VariBAD~\citep{zintgraf2019varibad} trains the recurrent hidden state to be predictive of the dynamics and rewards, which helps the hidden state approximate the belief state, a sufficient statistic of the history~\citep{kaelbling1998planning}, though gradients are not propagated fully end-to-end through the policy network.
\import~\citep{kamienny2020learning} regularizes the recurrent hidden state to be close to a learned embedding of the meta-training tasks IDs.

\textbf{\dream.}
While end-to-end approaches can learn the optimal policy in theory, they can run into optimization issues that make it difficult to learn sophisticated behaviors for the exploration episodes~\citep{liu2021dream}.
Specifically, from a high level, the exploration policy $\pi^\text{exp}$ is not incentivized to gather useful information, unless the task policy $\pi^\text{task}$ already can use that information.
However, $\pi^\text{task}$ cannot learn to use that information, unless $\pi^\text{exp}$ already gathers it.
\dream~\citep{liu2021dream} sidesteps this issue by training $\pi^\text{exp}$ on a separate objective, and we therefore primarily focus on \dream in our experiments (Section~\ref{sec:experiments}).

From a high level, \dream aims to create an objective for $\pi^\text{exp}$ to uncover exactly the information needed by the task policy $\pi^\text{task}$ to solve the task.
Doing this requires two key pieces:
(1) determining what information is needed by $\pi^\text{task}$;
and (2) creating an objective to recover this information.

\dream achieves (1) by training $\pi^\text{task}(a \mid s, z)$ to solve each task conditional on an encoding $z$ of the ID of the task $\mu$.
The idea is that if $\pi^\text{task}$ can solve each task conditioned on $z$, then $z$ contains all the information that it needs.
However, $z$ could also potentially contain extraneous information, which \dream attempts to remove by placing an information bottleneck~\citep{alemi2016deep} on it.
Once $z$ has been learned for all the meta-training tasks, \dream accomplishes (2) by proposing to maximize the mutual information $I(\tau^\text{exp}; z)$ between episodes $\tau^\text{exp}$ from the exploration policy $\pi^\text{exp}$ and $z$.
In other words, this objective trains $\pi^\text{exp}$ to recover exactly the information $z$ needed by $\pi^\text{task}$.

\section{Designing an Environment for Evaluating Language Emergence}
\label{sec:setup}

In this section, we detail our environment design for testing if RL agents can learn language as a byproduct of solving non-language tasks.
At a high level, answering this requires an environment meeting the previously mentioned criteria:
\begin{enumerate}[label=(\arabic*), leftmargin=*, itemsep=0pt, topsep=0pt]
    \item The environment must include language.
    \item Learning language must benefit the agent.
    \item Tasks must be solvable \emph{without} language understanding. Otherwise, any language learning is prescribed by the task, rather than emerging as a byproduct.
    \item Language should be observed in an unprocessed form, such as visually, rather than receiving string tokens.
\end{enumerate}

Note that simply adding static language into the environment, such as a sign with fixed text, is insufficient to encourage language learning to satisfy the second criteria.
Even if the sign contains useful information, an agent with no \emph{a priori} language understanding cannot learn what the sign means without observing its corresponding meaning in the world.
Further, once the agent has observed the meaning in the world, reading the sign now contains no useful new information.
Instead, language learning requires varied language whose meanings are reflected in the environment.

We therefore opt for a multi-task setting, where both the environment and the language within it change across tasks.
To enable the agent to adapt to new, but related tasks, we allow the agent a single ``free'' episode to explore, yielding the few-shot meta-RL setting in Section~\ref{sec:meta_rl_setting}.
We next detail our office environment and the language within it.

\subsection{The Office Environment}\label{sec:office_env}
To meet the above criteria in the few-shot meta-RL setting, we design an office environment.
The environment consists of several office rooms in a building, each of which is identified by one of six colors.
In each task, the office locations are randomized, and the agent must enter the blue office room,
but does not know where it is a priori.
We incorporate language into the environment with a \emph{floor plan} that describes the location of the blue office as either a written sign or a pictorial map, which can be \emph{visually} observed as an RGB array.
However, to ensure that language learning remains indirect, the agent receives no direct reward to incentivize viewing the floor plan.
We further describe these floor plans below.

This environment meets our criteria as follows.
The presence of visually perceived language floor plans satisfy criteria (1) and (4).
Crucially, it is possible to achieve optimal returns without learning language by exhaustively checking each room during the exploration episode to locate the blue goal office, satisfying criteria (3).
Recall that the objective (Equation~\ref{eqn:meta_rl_objective}) only depends on the returns achieved during the evaluation episode.
Hence, even though exhaustively visiting each room takes more timesteps during the exploration episode than visiting and reading the floor plan, it is an equally optimal strategy, as only the timesteps during the \emph{evaluation} episode affect the objective.
Additionally, learning to read the floor plans, though unnecessary, immediately reveals the blue office's location, allowing it to consistently navigate there as quickly as possible to maximize its returns, satisfying criteria (2).
Finally to robustly test reading comprehension, we design train / test splits in the environment that test generalization to held out language phrases, pictorial maps, and building layouts.

\begin{figure}[t]
\begin{center}
\centerline{\includegraphics[width=\columnwidth]{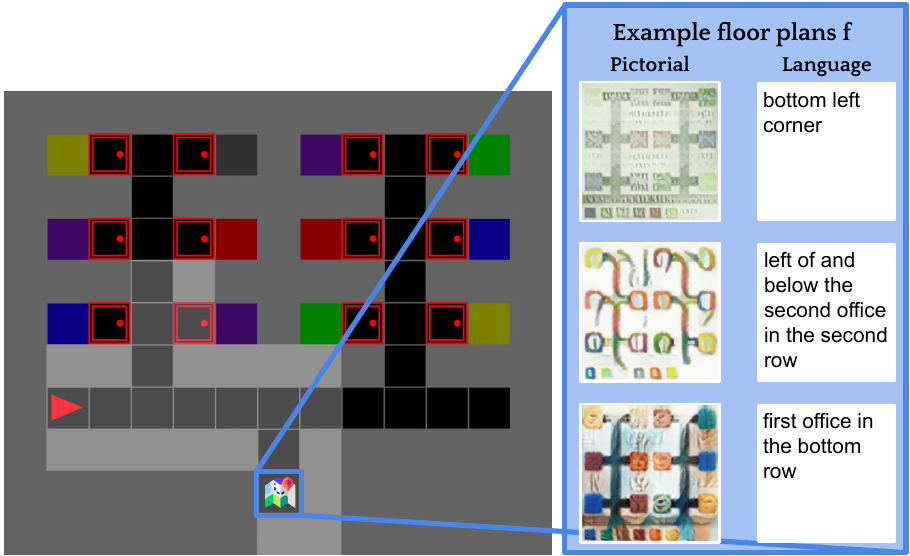}}
\vspace{-1mm}
\caption{The 2D variant.
The agent (red triangle) observes the highlighted cells in front of it and must visit the blue office.
Reading the floor plan in the bottom center gives the agent an RGB image like those on the right, which can help identify the location of the blue office. }
\vspace{-1mm}
\label{fig:2d}
\end{center}
\end{figure}

\textbf{Two-dimensional variant.}
We build two variants of this office environment: 2D and 3D.
The 2D environment contains low-dimensional observations to isolate the challenge of language learning, though the floor plans are still perceived as high-dimensional RGB images.
In contrast, the 3D environment tests scaling to purely high-dimensional pixel observations, and more closely resembles the real world.
We implement the 2D variant in Minigrid~\citep{minigrid} and visualize it in Figure~\ref{fig:2d}.
There are two hallways in this variant, each with six rooms.
In different tasks, the color identifiers of the 12 rooms are randomized.
Additionally, the agent can view the visual floor plan when located at the bottom center cell.

\textbf{Three-dimensional variant.}
The 3D variant is implemented in MiniWorld~\citep{gym_miniworld} and is visualized in Figure~\ref{fig:3d}.
In this variant, there are four office rooms identified by a colored block inside, and a floor plan opposite the office rooms that can be visually read by opening a door and walking to it.
The colors of the blocks are randomized in different tasks and the goal is to enter the office with the blue block.

In both versions, the environment defines a family of Markov decision processes with:
\begin{itemize}[itemsep=0pt,leftmargin=*]
    \item \emph{State space.}
    The state $s$ is an egocentric observation.
    In the 3D variant, this is a single $80 \times 60 \times 3$ RGB-array,
    whereas the state in the 2D variant $s = (o, f)$ consists of two components:
    First, a $7\times7\times3$ egocentric observation $o$, corresponding to the $7\times7$ grid cells in the agent's line of sight (the highlighted region in Figure~\ref{fig:2d}).
    Each grid cell is encoded as a $3$-dimensional vector based on what object is present in the cell, and is represented as zeros if the cell is occluded to the agent.
    Second, a $84 \times 84 \times 3$ RGB image floor plan observation $f$.
    This observation is an image representation of the floor plan when the agent is at the floor plan and otherwise is a blank image (all zeros).
    We detail the floor plan $f$ in the next section.
    \item \emph{Action space and dynamics.} The agent's actions are to turn left or right and move forward.
    Additionally, the agent can open or close a door.
    \item \emph{Reward function.} The agent receives $+1$ reward for visiting the blue office, which ends the episode.
    Otherwise, it receives $-0.1$ reward per timestep, to encourage it to visit the blue office as quickly as possible.
    The reward function does not directly incentivize learning language: there is no reward associated with visiting the floor plan.
\end{itemize}
\subsection{Floor Plans}\label{sec:floor_plan}
\begin{figure}[t]
\begin{center}
\centerline{\includegraphics[width=0.9\columnwidth]{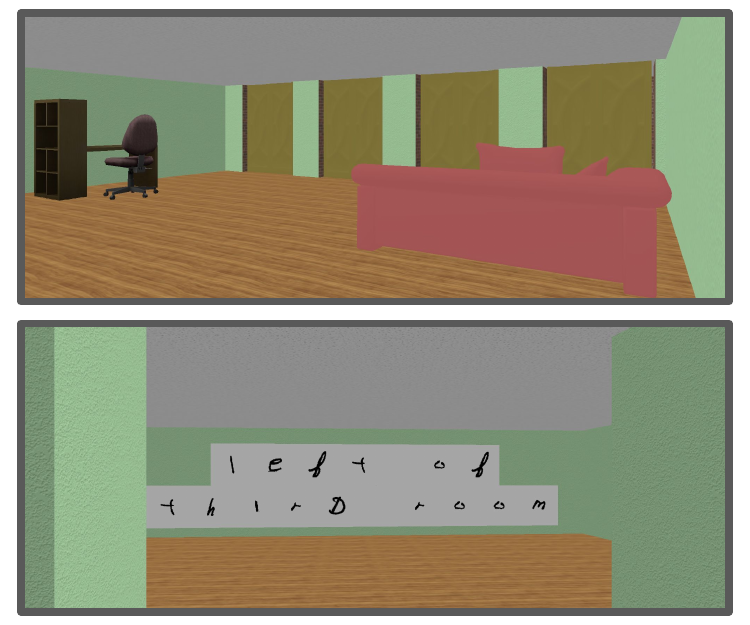}}
\vspace{-3mm}
\caption{The 3D office environment.
There are four offices and the agent must enter the correct one (above).
The agent can open a door to visually read a sign that states which office is correct (below), though it can also exhaustively check the four offices.}
\vspace{-3mm}
\label{fig:3d}
\end{center}
\end{figure}

In this section, we detail the floor plans that agents can read to determine the office locations.
These floor plans are readable by the agent via the observation $f$ in the 2D variant (Figure~\ref{fig:2d} right), and as a sign on the wall in the 3D variant (Figure~\ref{fig:3d} bottom).

\textbf{Language floor plan.}
We primarily focus on language floor plans, which provide text descriptions of where the goal room is.
Note that in both the 2D and 3D variants, this floor plan is still represented as a RGB image that the agent has to learn to parse---\emph{not} text tokens.
The details below use examples for the 2D variant, though the descriptions in the 3D variant are analogous, but modified for the 4 office layout.
The descriptions can either be direct (e.g.,  ``the second office in the third row'' or ``the top left corner'') or relative (e.g., ``right of the office right of the first office in the second row'').
Relative references can be chained one or more times, which we refer to as the \emph{relative step count} (e.g, ``above the office left of the \{\texttt{OFFICE LOCATION}\}'' has a relative step count of $2$).
In all of our experiments, unless otherwise noted, we restrict ourselves to relative step counts of $2$ or fewer.
Formally, descriptions $S$ are generated according to the following context-free grammar production rules:
{
\begin{align*}
S &\rightarrow \left(\text{top} \mid \text{bottom} \right) \left(\text{left} \mid \text{right} \right) \text{corner}, \\
S &\rightarrow \text{N office in the N row},  \\
S &\rightarrow \text{REL the } S, \\
REL &\rightarrow LR \mid AB \mid LR \text{ and } AB, \\
LR &\rightarrow \text{left of} \mid \text{right of}, \\
AB &\rightarrow \text{above} \mid \text{below}, \\
N &\rightarrow \text{first} \mid \text{second} \mid \text{third} \mid \text{fourth}. \\
\end{align*}
}\textbf{Pictorial floor plan.}
To test multiple modalities of reading, we additionally add a pictorial floor plan type, only to the 2D variant.
This type is a stylized RGB image of a fully observed top-down view of the building, which indicates the color identifier of each office room, and can be used to find the goal blue office.
We generate these by applying a style transfer model~\citep{gatys2016image}, where the content image is the unstylized top-down view of the environment, and the style image is a randomly sampled image from a book covers dataset~\citep{kaggle_bookcovers}.
Since the style transfer can recolor the offices, we also include a legend of the six possible colors in a fixed order to the bottom left of the content image.
The agent then can locate the blue office by finding offices that match the color with the first element in the legend.

\textbf{Demonstrations.}
To accelerate policy learning, without affecting the language learning, we pretrain only the task policy $\pi^\text{task}$ with demonstrations from a scripted policy that computes the shortest path between the agent and the goal blue office.
Critically, the exploration policy $\pi^\text{exp}$ does not receive any demonstrations to ensure that any language skills it acquires arise as a byproduct of meta-RL, rather than from imitating demonstrations.
We pretrain the task policy by including the demonstrations in its replay buffer.

\section{Experiments}
\label{sec:experiments}
In our experiments, we aim to answer four questions:
\begin{enumerate}[label=(\arabic*), leftmargin=*, itemsep=0pt, topsep=0pt]
    \item Our main question: Can agents learn language without explicit language supervision (Section~\ref{sec:language_exp})?
    \item Can agents learn to read other modalities beyond language, such as a pictorial map (Section~\ref{sec:picture_exp})?
    \item What factors impact language emergence (Section~\ref{sec:factors_exp})?
    \item Do these results scale to 3D environments with high-dimensional pixel observations (Section~\ref{sec:3d_exp})?
\end{enumerate}

\subsection{Language Emergence}\label{sec:language_exp}

\begin{figure*}[t]
\begin{center}
\centerline{\includegraphics[width=\textwidth]{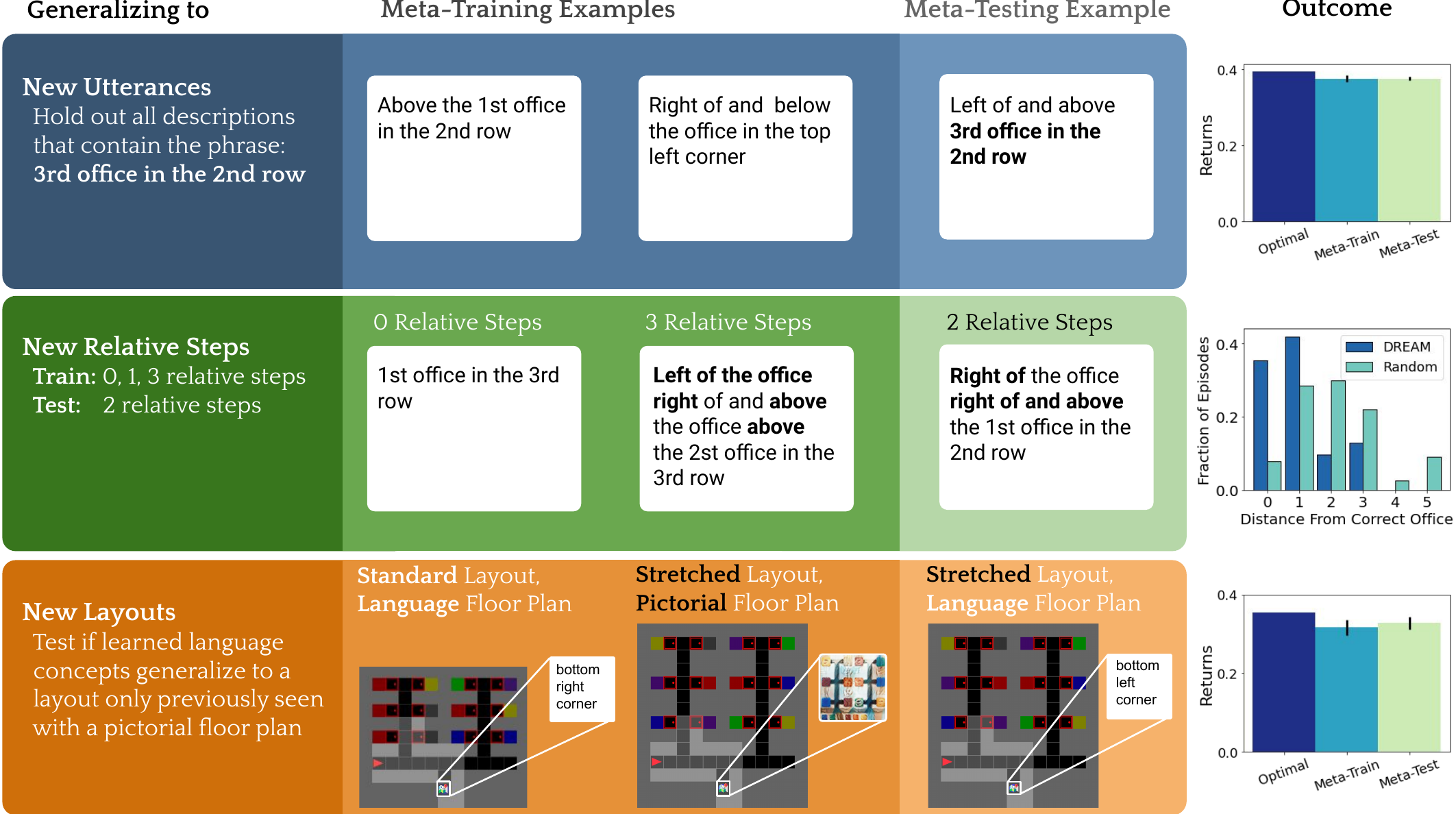}}
\vspace{-1mm}
\caption{Compositional generalization evaluations. 
\textbf{Top:} The agent generalizes to unseen descriptions, achieving similar returns on both training and held-out descriptions.
\textbf{Middle:} The agent also can partially generalize to unseen relative steps, where it almost always gets close to the correct office.
The right side plots the distribution of distances that the agent achieves from the correct office, as well as the distribution from randomly sampling an office.
The agent gets much closer to the correct office on average than sampling an office at random.
\textbf{Bottom:} The agent also can apply language concepts learned in one layout to a new layout, continuing to achieve near-optimal returns when evaluated on the stretched layout with language floor plans, which was not seen during training.
}
\vspace{-5mm}
\label{fig:language_results}
\end{center}
\end{figure*}

\begin{figure*}[t]
\begin{center}
\centerline{\includegraphics[width=\textwidth]{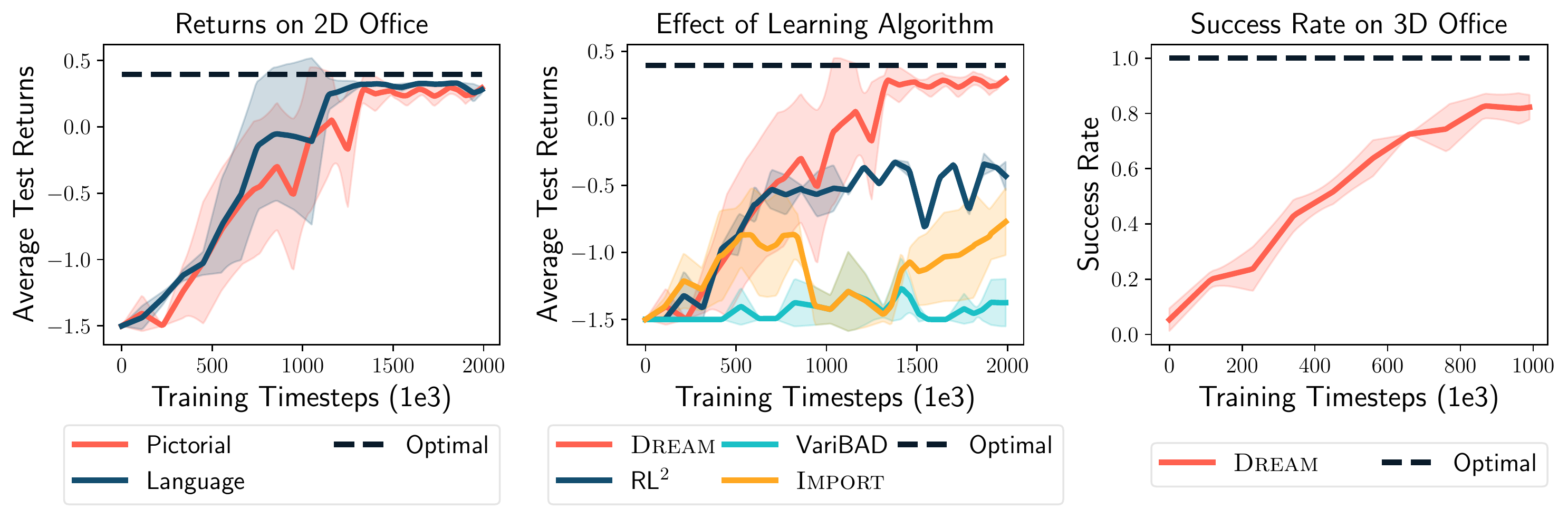}}
\vspace{-2mm}
\caption{Results averaged over 3 seeds, with 1-stddev error bars.
\textbf{Left:} \dream learns to read both pictorial and language floor plans and successfully uses information from the floor plan to achieve near-optimal returns.
\textbf{Middle:} Effect of learning algorithm on language learning. Only \dream learns to read and achieves near-optimal returns.
The other approaches check some, but not all of the offices, and consequently achieve lower returns.
\textbf{Right:} \dream learns to read the floor plan and solves the task in the 3D environment as well.
}
\vspace{-5mm}
\label{fig:combined_results}
\end{center}
\end{figure*}

To determine if language can emerge, we train \dream on the 2D office with language floor plans.
We begin in a simple setting where all language descriptions are seen during meta-training, and none are held out for meta-testing.

We find that \dream learns an exploration policy that navigates to and reads the floor plan.
Then, the task policy uses this information to directly walk to the goal office room, achieving near-optimal returns (Figure~\ref{fig:combined_results} left).
This result indicates that the agent has learned a form of simple language.
To understand any remaining failure modes, we qualitatively analyzed the trajectories where the agent did not achieve exactly optimal returns.
On these trajectories, the agent navigates to the floor plan during the exploration episode and correctly interprets the floor plan’s description to go to the correct office during the evaluation episodes.
However, during the evaluation episode, the agent takes a slightly circuitous route, such as using the “open door” action before reaching the office door. Since this takes more timesteps, the agent achieves slightly suboptimal returns.
Overall, the slight amount of suboptimality appears to only come from slight mistakes from the task policy, rather than any issues with reading and understanding the sign.

Next, to further understand the limits of the agent's learned language, we perform three tests of compositional generalization.

\textbf{Generalization to unseen descriptions.}
First, we test if the agent can generalize to new descriptions $S$ unseen during meta-training.
To achieve this, we hold out all language floor plans where ``3rd office in the 2nd row'' is a substring during meta-training and re-train \dream.
Note that we also hold out relative references, such as ``\emph{left of and above} the 3rd office in the 2nd row.''
This tests both if the agent can generalize to the held out direct reference ``3rd office in the 2nd row,'' as well as if it can correctly apply the prepositions ($LR$, $AB$) to it.
We find that the agent still reads the floor plan during exploration and successfully interprets the floor plans containing these held out descriptions during meta-testing to enter the correct office (Figure~\ref{fig:language_results}, top).

\textbf{Generalization to unseen relative step count.}
Next, we test if the agent can generalize to unseen relative step counts by training on floor plans with relative step counts of $0$, $1$ or $3$, and evaluating on floor plans with a relative step count of $2$.
Recall from Section~\ref{sec:setup} that the relative step count refers to the number of times relative references are chained.
For instance, ``left of the \{\texttt{OFFICE LOCATION}\}'' has a relative step count of $1$, while ``above the office left of the \{\texttt{OFFICE LOCATION}\}'' has a step count of $2$.
When presented with an unseen relative step count, we find that the agent successfully navigates to the correct office only around 37\% of the time.
However, the agent still displays some level of generalization.
Rather than randomly randomly visiting offices, the agent tends to navigate to an office nearby the correct office, suggesting that is successfully resolving some, but not all of the chained relative references.
Specifically, we find that the agent navigates to the correct office or a neighboring office 80\% of the time.
Furthermore, it gets much closer to the correct office on average than selecting an office at random (Figure~\ref{fig:language_results} middle).

\textbf{Generalizing to new layouts.}
Finally, we test if the agent's understanding of language can generalize to a new layout.
This helps us understand if the agent learns reusable concepts that can be applied in different situations, as opposed to memorizing an exact goal location for each description.
To achieve this, we create a new \emph{stretched} layout that doubles the distances between contiguous offices.
We allow the agent to familiarize itself with this stretched layout by training it with pictorial floor plans on the stretched layout, while still also training on the standard layout with language floor plans.
However, the agent never sees the stretched layout with a language floor plan during training.
Then, we test if the agent can generalize to language floor plans on the stretched layout at meta-test time.

Figure~\ref{fig:language_results} shows the results on the bottom.
At test time, we find that the agent still reads the language floor plan in the stretched layout, and further successfully interprets the result to go to the correct office nearly 100\% of the time.
This indicates that the agent has learned generalizable language concepts that it can apply to simple layout changes.

\textbf{Probing the learned representation of the floor plan.}
Beyond testing the agent's ability to compositionally generalize, we probe the the agent's learned representation of the language floor plan to test if it understands the floor plan by reading individual words, as opposed to leveraging spurious correlations (e.g., the length of the description).
We consider a probing task of extracting the tokens from the floor plan, i.e., extracting the string of the floor plan description from the image.
Specifically, we first freeze the agent's learned representation of the floor plan after we train it to solve the office environment. We then train a single layer LSTM~\citep{hochreiter1997lstm} on top of this frozen representation.

We train this probing layer on 80\% of the language floor plans, and evaluate it on a held out 20\% with the standard perplexity metric (i.e., the negative log likelihood of the correct tokens under the probing layer, where lower is better).
We compare the agent's frozen learned representation to a representation extracted from a randomly initialized network of the same architecture, and find that the agent's frozen representation achieves significantly lower perplexity.
Specifically, learning with the frozen representation achieves a perplexity of 1.1, while the random representation achieves a perplexity of 4.65.
This suggests that the agent indeed reads individual words of the floor plan, and can generally extract words from images.

\subsection{Learning to Read Other Modalities}\label{sec:picture_exp}
Given that the \dream-based agent successfully learns to read language in the above experiments, we ask if the agent can also read other modalities.
To test this, we train \dream on the 2D variant of the office with pictorial floor plans, instead of language floor plans.
We generate 100 floor plan types with a pre-trained style transfer model (Section~\ref{sec:floor_plan}), split into 90 training and 10 held out for testing.

Similar to the language floor plan case,
we find that \dream learns a policy that navigates to and reads the floor plan during its exploration episodes, and then directly enters the correct office in its evaluation episodes nearly 100\% of the time. 
Crucially, the agent successfully reads and interprets unseen pictorial floor plan styles at test time (Figure~\ref{fig:combined_results} left).

\subsection{What Factors Impact the Emergence of Reading?}\label{sec:factors_exp}
In this section, we evaluate the factors that may influence whether the agent learns to read.
Specifically, we study the impact varying parameters of the learning process and the environment:
\begin{itemize}[topsep=0pt]
    \item \emph{Learning algorithm.} We replace \dream with other state-of-the-art meta-RL algorithms.
    \item \emph{Amount of meta-training data.} We vary both the number of tasks (i.e., office color randomizations) and unique floor plans seen during meta-training.
    \item \emph{Size of the model.} We vary the number of final fully connected layers in \dream's policy network.
\end{itemize}

\textbf{Learning algorithm.}
We compare \dream with other state-of-the-art meta-RL algorithms on our 2D office environment with pictorial floor plans, namely \rl~\citep{duan2016rl,wang2016learning}, VariBAD~\citep{zintgraf2019varibad}, and \import~\citep{kamienny2020learning}.
See Appendix~\ref{sec:algorithm_details} for details and hyperparameters.
Figure~\ref{fig:combined_results} shows the results in the middle.
We find that the choice of meta-RL algorithm drastically affects whether the agent learns to read, and even whether the agent solves the task.
In contrast to \dream, the other algorithms do not learn language, and instead learn to check some of the other offices during the exploration episodes.
However, as they do not learn to check \emph{all} of the offices, they often fail to enter the correct office, achieving much lower returns than \dream.

\textbf{Amount of meta-training data.}
To study the effect of varying the amount of meta-training data, we train \dream on pictorial floor plans on the 2D variant of the office environment.
We vary the number of floor plans seen during meta-training ($30$, $60$, and $90$), and vary the number of meta-training tasks ($5$, $10$, $50$, $100$, $200$, and $500$).

We find that the number of meta-training tasks has a sizeable impact on whether the agent learns to read.
The agent continues to visit and read the floor plan when the number of meta-training tasks exceeds $5$, but instead switches to a strategy of checking the offices when the number of meta-training tasks falls to $5$---since there are only a few meta-training tasks, visiting only a few of the office rooms is sufficient to disambiguate them, and determine the location of the blue office.
In contrast, when there are many meta-training tasks, disambiguating the tasks requires visiting all or nearly all of the office rooms.
This result suggests that the agent learns to read when the information value of reading is high, and does not when there are easier means to obtain the same information.
Notably, the strategy of checking only a few office rooms is sufficient to disambiguate the meta-training tasks when there are only a few tasks, but it fails to generalize to new unseen tasks.

We find that the agent still learns to visit and read the floor plan regardless of the number of meta-training floor plans,
though, unsurprisingly, the agent's ability to generalize to new floor plans degrades as the number of floor plans decreases.

\textbf{Size of the model.}
Motivated by the observation that increasing model size can yield significant new capabilities in language models~\citep{wei2022emergent}, we study the effect of the policy network's model size.
We add one, two, or three additional fully connected
layers with ReLU activations to \dream's policy network in the pictorial floor plan version of the 2D office environment and tune over a grid of 3 layer widths (64, 128, and 256 hidden units).
We find that the agent continues to learn to read and solve the task, and further that with appropriate tuning, adding any number of additional layers (one, two or three) improves sample complexity by roughly 8\%.
However, we hypothesize that these improvements likely arise simply from the additional tuning over layer width, as the original results without added layers use the default \dream hyperparameters without tuning.

\subsection{High-Dimensional 3D World}\label{sec:3d_exp}
Our experiments thus far isolate the challenge of indirectly learning to read by leveraging a 2D domain.
In this section, we test if indirect language learning still occurs when we scale to a 3D domain with first-person pixel observations.
We find that training \dream on our 3D domain still indirectly learns to read.
The agent navigates to and reads the sign during its exploration episodes and then directly enters the correct office room during most evaluation episodes
(Figure~\ref{fig:combined_results} right).

\section{Conclusion}
\label{sec:conclusion}

In this work, we found that basic language can emerge as a byproduct of solving non-language tasks in meta-RL agents.
This previously unexplored paradigm for language learning offers potential, but not yet realized, benefits over standard language learning.
Specifically, one insidious and pervasive failure mode of standard large language models trained on datasets mined from the Internet is that they frequently output convincing and fluent, yet wildly incorrect statements~\citep{lin2021truthful}.
Indirectly learned language might avoid this failure mode, as the learned language is directly tied to observations and experiences in the world.
Additionally, language understanding indirectly gained from embodied experience can potentially be \emph{deeper} and yield greater physical intuitions.
For example, no amount of reading about sourness on the Internet can give as rich an understanding as actually \emph{experiencing} the sensation of biting a lemon.

However, we emphasize that such benefits are far from being realized.
Our work focused on relatively simple language on only a single domain of office navigation to illustrate a proof of concept.
To even begin to compare to the capabilities of standard language learning, the emergent learning paradigm requires significant future work with more sophisticated environments and language. 
Furthermore, betting against data has been a losing proposition in machine learning~\citep{sutton2019bitter}.
Even so, considering alternatives to even powerful existing approaches is a key piece to driving scientific progress---and perhaps the best way forward is to combine the approaches by augmenting emergent language learning with direct language supervision.

\section*{Acknowledgements}

We thank Annie Xie and other members of the IRIS lab for helpful feedback on initial drafts of this work.
We thank Kyle Hsu for providing access to the ShapeNet dataset used in the 3D office variant.
We thank Eric Mitchell for his expert opinions on natural language processing.

This work is supported in part by Intel.
CF is a CIFAR fellow in the Learning in Machines and Brains program.
EZL is supported by a National Science Foundation Graduate Research Fellowship under Grant No.
DGE-1656518.
Icons in the figures were made by JungleOutThere and ilyakalinin from Adobe Stock.

\bibliography{all}

\begin{thebibliography}{49}
\providecommand{\natexlab}[1]{#1}
\providecommand{\url}[1]{\texttt{#1}}
\expandafter\ifx\csname urlstyle\endcsname\relax
  \providecommand{\doi}[1]{doi: #1}\else
  \providecommand{\doi}{doi: \begingroup \urlstyle{rm}\Url}\fi

\bibitem[Ahn et~al.(2022)Ahn, Brohan, Brown, Chebotar, Cortes, David, Finn,
  Gopalakrishnan, Hausman, Herzog, Ho, Hsu, Ibarz, Ichter, Irpan, Jang, Ruano,
  Jeffrey, Jesmonth, Joshi, Julian, Kalashnikov, Kuang, Lee, Levine, Lu, Luu,
  Parada, Pastor, Quiambao, Rao, Rettinghouse, Reyes, Sermanet, Sievers, Tan,
  Toshev, Vanhoucke, Xia, Xiao, Xu, Xu, and Yan]{ahn2022saycan}
Ahn, M., Brohan, A., Brown, N., Chebotar, Y., Cortes, O., David, B., Finn, C.,
  Gopalakrishnan, K., Hausman, K., Herzog, A., Ho, D., Hsu, J., Ibarz, J.,
  Ichter, B., Irpan, A., Jang, E., Ruano, R.~J., Jeffrey, K., Jesmonth, S.,
  Joshi, N.~J., Julian, R.~C., Kalashnikov, D., Kuang, Y., Lee, K.-H., Levine,
  S., Lu, Y., Luu, L., Parada, C., Pastor, P., Quiambao, J., Rao, K.,
  Rettinghouse, J., Reyes, D.~M., Sermanet, P., Sievers, N., Tan, C., Toshev,
  A., Vanhoucke, V., Xia, F., Xiao, T., Xu, P., Xu, S., and Yan, M.
\newblock Do as {I} can, not as {I} say: Grounding language in robotic
  affordances.
\newblock \emph{arXiv preprint arXiv:2204.01691}, 2022.

\bibitem[Alemi et~al.(2016)Alemi, Fischer, Dillon, and Murphy]{alemi2016deep}
Alemi, A.~A., Fischer, I., Dillon, J.~V., and Murphy, K.
\newblock Deep variational information bottleneck.
\newblock \emph{arXiv preprint arXiv:1612.00410}, 2016.

\bibitem[Andreas et~al.(2017)Andreas, Klein, and Levine]{andreas2017learning}
Andreas, J., Klein, D., and Levine, S.
\newblock Learning with latent language.
\newblock \emph{arXiv preprint arXiv:1711.00482}, 2017.

\bibitem[Anicin(2020)]{kaggle_bookcovers}
Anicin, L.
\newblock Book covers dataset (kaggle).
\newblock \url{https://www.kaggle.com/datasets/lukaanicin/book-covers-dataset},
  2020.

\bibitem[Baker et~al.(2019)Baker, Kanitscheider, Markov, Wu, Powell, McGrew,
  and Mordatch]{baker2019emergent}
Baker, B., Kanitscheider, I., Markov, T., Wu, Y., Powell, G., McGrew, B., and
  Mordatch, I.
\newblock Emergent tool use from multi-agent autocurricula.
\newblock \emph{arXiv preprint arXiv:1909.07528}, 2019.

\bibitem[Bansal et~al.(2017)Bansal, Pachocki, Sidor, Sutskever, and
  Mordatch]{bansal2017emergent}
Bansal, T., Pachocki, J., Sidor, S., Sutskever, I., and Mordatch, I.
\newblock Emergent complexity via multi-agent competition.
\newblock \emph{arXiv preprint arXiv:1710.03748}, 2017.

\bibitem[Brown et~al.(2020)Brown, Mann, Ryder, Subbiah, Kaplan, Dhariwal,
  Neelakantan, Shyam, Sastry, Askell, Agarwal, Herbert-Voss, Krueger, Henighan,
  Child, Ramesh, Ziegler, Wu, Winter, Hesse, Chen, Sigler, Litwin, Gray, Chess,
  Clark, Berner, McCandlish, Radford, Sutskever, and Amodei]{brown2020gpt3}
Brown, T.~B., Mann, B., Ryder, N., Subbiah, M., Kaplan, J., Dhariwal, P.,
  Neelakantan, A., Shyam, P., Sastry, G., Askell, A., Agarwal, S.,
  Herbert-Voss, A., Krueger, G., Henighan, T., Child, R., Ramesh, A., Ziegler,
  D.~M., Wu, J., Winter, C., Hesse, C., Chen, M., Sigler, E., Litwin, M., Gray,
  S., Chess, B., Clark, J., Berner, C., McCandlish, S., Radford, A., Sutskever,
  I., and Amodei, D.
\newblock Language models are few-shot learners.
\newblock \emph{arXiv preprint arXiv:2005.14165}, 2020.

\bibitem[Chevalier-Boisvert(2018)]{gym_miniworld}
Chevalier-Boisvert, M.
\newblock Miniworld: Minimalistic 3d environment for rl \& robotics research.
\newblock \url{https://github.com/maximecb/gym-miniworld}, 2018.

\bibitem[Chevalier-Boisvert et~al.(2018)Chevalier-Boisvert, Willems, and
  Pal]{minigrid}
Chevalier-Boisvert, M., Willems, L., and Pal, S.
\newblock Minimalistic gridworld environment for gymnasium, 2018.
\newblock URL \url{https://github.com/Farama-Foundation/Minigrid}.

\bibitem[Devlin et~al.(2019)Devlin, Chang, Lee, and Toutanova]{devlin2019bert}
Devlin, J., Chang, M.-W., Lee, K., and Toutanova, K.
\newblock {BERT}: Pre-training of deep bidirectional transformers for language
  understanding.
\newblock In \emph{Association for Computational Linguistics (ACL)}, pp.\
  4171--4186, 2019.

\bibitem[Duan et~al.(2016)Duan, Schulman, Chen, Bartlett, Sutskever, and
  Abbeel]{duan2016rl}
Duan, Y., Schulman, J., Chen, X., Bartlett, P.~L., Sutskever, I., and Abbeel,
  P.
\newblock {RL}$^2$: Fast reinforcement learning via slow reinforcement
  learning.
\newblock \emph{arXiv preprint arXiv:1611.02779}, 2016.

\bibitem[Finn et~al.(2017)Finn, Abbeel, and Levine]{finn2017modelagnostic}
Finn, C., Abbeel, P., and Levine, S.
\newblock Model-agnostic meta-learning for fast adaptation of deep networks.
\newblock In \emph{International Conference on Machine Learning (ICML)}, 2017.

\bibitem[Fu et~al.(2019)Fu, Korattikara, Levine, and
  Guadarrama]{fu2019language}
Fu, J., Korattikara, A., Levine, S., and Guadarrama, S.
\newblock From language to goals: Inverse reinforcement learning for
  vision-based instruction following.
\newblock \emph{arXiv preprint arXiv:1902.07742}, 2019.

\bibitem[Gatys et~al.(2016)Gatys, Ecker, and Bethge]{gatys2016image}
Gatys, L.~A., Ecker, A.~S., and Bethge, M.
\newblock Image style transfer using convolutional neural networks.
\newblock In \emph{Proceedings of the IEEE conference on computer vision and
  pattern recognition}, pp.\  2414--2423, 2016.

\bibitem[Hermann et~al.(2017)Hermann, Hill, Green, Wang, Faulkner, Soyer,
  Szepesvari, Czarnecki, Jaderberg, Teplyashin, Wainwright, Apps, Hassabis, and
  Blunsom]{hermann2017grounded}
Hermann, K.~M., Hill, F., Green, S., Wang, F., Faulkner, R., Soyer, H.,
  Szepesvari, D., Czarnecki, W., Jaderberg, M., Teplyashin, D., Wainwright, M.,
  Apps, C., Hassabis, D., and Blunsom, P.
\newblock Grounded language learning in a simulated 3d world.
\newblock \emph{arXiv preprint arXiv:1706.06551}, 2017.

\bibitem[Hill et~al.(2020{\natexlab{a}})Hill, Mokra, Wong, and
  Harley]{hill2020human}
Hill, F., Mokra, S., Wong, N., and Harley, T.
\newblock Human instruction-following with deep reinforcement learning via
  transfer-learning from text.
\newblock \emph{arXiv preprint arXiv:2005.09382}, 2020{\natexlab{a}}.

\bibitem[Hill et~al.(2020{\natexlab{b}})Hill, Tieleman, Von~Glehn, Wong,
  Merzic, and Clark]{hill2020grounded}
Hill, F., Tieleman, O., Von~Glehn, T., Wong, N., Merzic, H., and Clark, S.
\newblock Grounded language learning fast and slow.
\newblock \emph{arXiv preprint arXiv:2009.01719}, 2020{\natexlab{b}}.

\bibitem[Hochreiter \& Schmidhuber(1997)Hochreiter and
  Schmidhuber]{hochreiter1997lstm}
Hochreiter, S. and Schmidhuber, J.
\newblock Long short-term memory.
\newblock \emph{Neural Computation}, 9\penalty0 (8):\penalty0 1735--1780, 1997.

\bibitem[Jaderberg et~al.(2019)Jaderberg, Czarnecki, Dunning, Marris, Lever,
  Castaneda, Beattie, Rabinowitz, Morcos, Ruderman, et~al.]{jaderberg2019human}
Jaderberg, M., Czarnecki, W.~M., Dunning, I., Marris, L., Lever, G., Castaneda,
  A.~G., Beattie, C., Rabinowitz, N.~C., Morcos, A.~S., Ruderman, A., et~al.
\newblock Human-level performance in 3d multiplayer games with population-based
  reinforcement learning.
\newblock \emph{Science}, 364\penalty0 (6443):\penalty0 859--865, 2019.

\bibitem[Jiang et~al.(2019)Jiang, Gu, Murphy, and Finn]{jiang2019abstraction}
Jiang, Y., Gu, S.~S., Murphy, K.~P., and Finn, C.
\newblock Language as an abstraction for hierarchical deep reinforcement
  learning.
\newblock In \emph{Advances in Neural Information Processing Systems
  (NeurIPS)}, 2019.

\bibitem[Kaelbling et~al.(1998)Kaelbling, Littman, and
  Cassandra]{kaelbling1998planning}
Kaelbling, L.~P., Littman, M.~L., and Cassandra, A.~R.
\newblock Planning and acting in partially observable stochastic domains.
\newblock \emph{Artificial intelligence}, 101\penalty0 (1):\penalty0 99--134,
  1998.

\bibitem[Kamienny et~al.(2020)Kamienny, Pirotta, Lazaric, Lavril, Usunier, and
  Denoyer]{kamienny2020learning}
Kamienny, P.-A., Pirotta, M., Lazaric, A., Lavril, T., Usunier, N., and
  Denoyer, L.
\newblock Learning adaptive exploration strategies in dynamic environments
  through informed policy regularization.
\newblock \emph{arXiv preprint arXiv:2005.02934}, 2020.

\bibitem[Kreutzer et~al.(2020)Kreutzer, Riezler, and
  Lawrence]{kreutzer2020offline}
Kreutzer, J., Riezler, S., and Lawrence, C.
\newblock Offline reinforcement learning from human feedback in real-world
  sequence-to-sequence tasks.
\newblock \emph{arXiv preprint arXiv:2011.02511}, 2020.

\bibitem[K{\"u}ttler et~al.(2020)K{\"u}ttler, Nardelli, Miller, Raileanu,
  Selvatici, Grefenstette, and Rockt{\"a}schel]{kuttler2020nethack}
K{\"u}ttler, H., Nardelli, N., Miller, A., Raileanu, R., Selvatici, M.,
  Grefenstette, E., and Rockt{\"a}schel, T.
\newblock The nethack learning environment.
\newblock \emph{Advances in Neural Information Processing Systems},
  33:\penalty0 7671--7684, 2020.

\bibitem[Li et~al.(2020)Li, Sun, Han, and Li]{li2020survey}
Li, J., Sun, A., Han, J., and Li, C.
\newblock A survey on deep learning for named entity recognition.
\newblock \emph{IEEE Transactions on Knowledge and Data Engineering},
  34\penalty0 (1):\penalty0 50--70, 2020.

\bibitem[Lin et~al.(2021{\natexlab{a}})Lin, Hilton, and Evans]{lin2021truthful}
Lin, S., Hilton, J., and Evans, O.
\newblock Truthfulqa: Measuring how models mimic human falsehoods.
\newblock \emph{arXiv preprint arXiv:2109.07958}, 2021{\natexlab{a}}.

\bibitem[Lin et~al.(2021{\natexlab{b}})Lin, Hilton, and
  Evans]{lin2021truthfulqa}
Lin, S., Hilton, J., and Evans, O.
\newblock Truthfulqa: Measuring how models mimic human falsehoods.
\newblock \emph{arXiv preprint arXiv:2109.07958}, 2021{\natexlab{b}}.

\bibitem[Ling \& Fidler(2017)Ling and Fidler]{ling2017teaching}
Ling, H. and Fidler, S.
\newblock Teaching machines to describe images via natural language feedback.
\newblock In \emph{Advances in Neural Information Processing Systems
  (NeurIPS)}, 2017.

\bibitem[Liu et~al.(2021)Liu, Raghunathan, Liang, and Finn]{liu2021dream}
Liu, E.~Z., Raghunathan, A., Liang, P., and Finn, C.
\newblock Decoupling exploration and exploitation for meta-reinforcement
  learning without sacrifices.
\newblock In \emph{International Conference on Machine Learning (ICML)}, 2021.

\bibitem[Luketina et~al.(2019)Luketina, Nardelli, Farquhar, Foerster, Andreas,
  Grefenstette, Whiteson, and Rockt{\"{a}}schel]{luketina2019survey}
Luketina, J., Nardelli, N., Farquhar, G., Foerster, J., Andreas, J.,
  Grefenstette, E., Whiteson, S., and Rockt{\"{a}}schel, T.
\newblock A survey of reinforcement learning informed by natural language.
\newblock In \emph{International Joint Conference on Artificial Intelligence
  (IJCAI)}, 2019.

\bibitem[Misra et~al.(2017)Misra, Langford, and Artzi]{misra2017mapping}
Misra, D.~K., Langford, J., and Artzi, Y.
\newblock Mapping instructions and visual observations to actions with
  reinforcement learning.
\newblock In \emph{Empirical Methods in Natural Language Processing (EMNLP)},
  2017.

\bibitem[Mnih et~al.(2015)Mnih, Kavukcuoglu, Silver, Rusu, Veness, Bellemare,
  Graves, Riedmiller, Fidjeland, Ostrovski, et~al.]{mnih2015human}
Mnih, V., Kavukcuoglu, K., Silver, D., Rusu, A.~A., Veness, J., Bellemare,
  M.~G., Graves, A., Riedmiller, M., Fidjeland, A.~K., Ostrovski, G., et~al.
\newblock Human-level control through deep reinforcement learning.
\newblock \emph{Nature}, 518\penalty0 (7540):\penalty0 529--533, 2015.

\bibitem[Narasimhan et~al.(2015)Narasimhan, Kulkarni, and
  Barzilay]{narasimhan2015language}
Narasimhan, K., Kulkarni, T., and Barzilay, R.
\newblock Language understanding for text-based games using deep reinforcement
  learning.
\newblock \emph{arXiv preprint arXiv:1506.08941}, 2015.

\bibitem[Paszke et~al.(2019)Paszke, Gross, Massa, Lerer, Bradbury, Chanan,
  Killeen, Lin, Gimelshein, Antiga, Desmaison, K{\"o}pf, Yang, DeVito, Raison,
  Tejani, Chilamkurthy, Steiner, Fang, Bai, and Chintala]{paszke2019pytorch}
Paszke, A., Gross, S., Massa, F., Lerer, A., Bradbury, J., Chanan, G., Killeen,
  T., Lin, Z., Gimelshein, N., Antiga, L., Desmaison, A., K{\"o}pf, A., Yang,
  E., DeVito, Z., Raison, M., Tejani, A., Chilamkurthy, S., Steiner, B., Fang,
  L., Bai, J., and Chintala, S.
\newblock Pytorch: An imperative style, high-performance deep learning library.
\newblock In \emph{Advances in Neural Information Processing Systems
  (NeurIPS)}, 2019.

\bibitem[Radford et~al.(2018)Radford, Narasimhan, Salimans, and
  Sutskever]{radford2018improving}
Radford, A., Narasimhan, K., Salimans, T., and Sutskever, I.
\newblock Improving language understanding by generative pre-training.
\newblock Technical report, OpenAI, 2018.

\bibitem[Shah et~al.(2018)Shah, Fiser, Faust, Kew, and
  Hakkani-Tur]{shah2018follownet}
Shah, P., Fiser, M., Faust, A., Kew, J.~C., and Hakkani-Tur, D.
\newblock Follownet: Robot navigation by following natural language directions
  with deep reinforcement learning.
\newblock \emph{arXiv preprint arXiv:1805.06150}, 2018.

\bibitem[Sumers et~al.(2021)Sumers, Ho, Hawkins, Narasimhan, and
  Griffiths]{sumers2021learning}
Sumers, T.~R., Ho, M.~K., Hawkins, R.~D., Narasimhan, K., and Griffiths, T.~L.
\newblock Learning rewards from linguistic feedback.
\newblock In \emph{Proceedings of the AAAI Conference on Artificial
  Intelligence}, volume~35, pp.\  6002--6010, 2021.

\bibitem[Sutton(2019)]{sutton2019bitter}
Sutton, R.
\newblock The bitter lesson.
\newblock \emph{Incomplete Ideas (blog)}, 13\penalty0 (1), 2019.

\bibitem[Team et~al.(2023)Team, Bauer, Baumli, Baveja, Behbahani, Bhoopchand,
  Bradley-Schmieg, Chang, Clay, Collister, et~al.]{team2023human}
Team, A.~A., Bauer, J., Baumli, K., Baveja, S., Behbahani, F., Bhoopchand, A.,
  Bradley-Schmieg, N., Chang, M., Clay, N., Collister, A., et~al.
\newblock Human-timescale adaptation in an open-ended task space.
\newblock \emph{arXiv preprint arXiv:2301.07608}, 2023.

\bibitem[Team et~al.(2021)Team, Stooke, Mahajan, Barros, Deck, Bauer,
  Sygnowski, Trebacz, Jaderberg, Mathieu, et~al.]{team2021open}
Team, O. E.~L., Stooke, A., Mahajan, A., Barros, C., Deck, C., Bauer, J.,
  Sygnowski, J., Trebacz, M., Jaderberg, M., Mathieu, M., et~al.
\newblock Open-ended learning leads to generally capable agents.
\newblock \emph{arXiv preprint arXiv:2107.12808}, 2021.

\bibitem[Vaezipoor et~al.(2021)Vaezipoor, Li, Icarte, and
  Mcilraith]{vaezipoor2021ltl2action}
Vaezipoor, P., Li, A.~C., Icarte, R. A.~T., and Mcilraith, S.~A.
\newblock Ltl2action: Generalizing ltl instructions for multi-task rl.
\newblock In \emph{International Conference on Machine Learning}, pp.\
  10497--10508. PMLR, 2021.

\bibitem[Vaswani et~al.(2017)Vaswani, Shazeer, Parmar, Uszkoreit, Jones, Gomez,
  Kaiser, and Polosukhin]{vaswani2017attention}
Vaswani, A., Shazeer, N., Parmar, N., Uszkoreit, J., Jones, L., Gomez, A.~N.,
  Kaiser, L., and Polosukhin, I.
\newblock Attention is all you need.
\newblock \emph{arXiv preprint arXiv:1706.03762}, 2017.

\bibitem[Wang et~al.(2016)Wang, Kurth-Nelson, Tirumala, Soyer, Leibo, Munos,
  Blundell, Kumaran, and Botvinick]{wang2016learning}
Wang, J.~X., Kurth-Nelson, Z., Tirumala, D., Soyer, H., Leibo, J.~Z., Munos,
  R., Blundell, C., Kumaran, D., and Botvinick, M.
\newblock Learning to reinforcement learn.
\newblock \emph{arXiv preprint arXiv:1611.05763}, 2016.

\bibitem[Wei et~al.(2022)Wei, Tay, Bommasani, Raffel, Zoph, Borgeaud, Yogatama,
  Bosma, Zhou, Metzler, Chi, Hashimoto, Vinyals, Liang, Dean, and
  Fedus]{wei2022emergent}
Wei, J., Tay, Y., Bommasani, R., Raffel, C., Zoph, B., Borgeaud, S., Yogatama,
  D., Bosma, M., Zhou, D., Metzler, D., Chi, E., Hashimoto, T., Vinyals, O.,
  Liang, P., Dean, J., and Fedus, W.
\newblock Emergent abilities of large language models.
\newblock \emph{Transcations of Machine Learning Research (TMLR)}, 0, 2022.

\bibitem[Yan et~al.(2022)Yan, Carnevale, Georgiev, Santoro, Guy, Muldal, Hung,
  Abramson, Lillicrap, and Wayne]{yan2022intra}
Yan, C., Carnevale, F., Georgiev, P., Santoro, A., Guy, A., Muldal, A., Hung,
  C.-C., Abramson, J., Lillicrap, T., and Wayne, G.
\newblock Intra-agent speech permits zero-shot task acquisition.
\newblock \emph{arXiv preprint arXiv:2206.03139}, 2022.

\bibitem[Yuan et~al.(2018)Yuan, C{\^o}t{\'e}, Sordoni, Laroche, Combes,
  Hausknecht, and Trischler]{yuan2018counting}
Yuan, X., C{\^o}t{\'e}, M.-A., Sordoni, A., Laroche, R., Combes, R. T.~d.,
  Hausknecht, M., and Trischler, A.
\newblock Counting to explore and generalize in text-based games.
\newblock \emph{arXiv preprint arXiv:1806.11525}, 2018.

\bibitem[Zhang et~al.(2018)Zhang, Wang, and Liu]{zhang2018deep}
Zhang, L., Wang, S., and Liu, B.
\newblock Deep learning for sentiment analysis: A survey.
\newblock \emph{Wiley Interdisciplinary Reviews: Data Mining and Knowledge
  Discovery}, 8\penalty0 (4):\penalty0 e1253, 2018.

\bibitem[Zheng et~al.(2020)Zheng, Trott, Srinivasa, Naik, Gruesbeck, Parkes,
  and Socher]{zheng2020ai}
Zheng, S., Trott, A., Srinivasa, S., Naik, N., Gruesbeck, M., Parkes, D.~C.,
  and Socher, R.
\newblock The ai economist: Improving equality and productivity with ai-driven
  tax policies.
\newblock \emph{arXiv preprint arXiv:2004.13332}, 2020.

\bibitem[Zintgraf et~al.(2019)Zintgraf, Shiarlis, Igl, Schulze, Gal, Hofmann,
  and Whiteson]{zintgraf2019varibad}
Zintgraf, L., Shiarlis, K., Igl, M., Schulze, S., Gal, Y., Hofmann, K., and
  Whiteson, S.
\newblock Varibad: A very good method for bayes-adaptive deep {RL} via
  meta-learning.
\newblock \emph{arXiv preprint arXiv:1910.08348}, 2019.

\end{thebibliography}
\bibliographystyle{icml2023}

\newpage

\appendix
\onecolumn
\section{Experimental Details}\label{sec:algorithm_details}

We build upon the PyTorch~\citep{paszke2019pytorch} \dream, \import, \rl and VariBAD implementations released by \citet{liu2021dream}.

\subsection{Model Architecture}

For all methods, we use the default architectures detailed in Appendix B.3 of \citet{liu2021dream} with custom state embedders $e(s)$.

\textbf{Two-dimensional office variant.} 
Recall that the state $s = (o, f)$ in the 2D variant consists of two components, the egocentric observation $o$ and the map observation $f$.
We compute embeddings of the state $e(s)$ by first computing embeddings $e(o)$ and $e(f)$ of the two components, and then applying two final fully-connected layers of output size 64 with a ReLU activation.
We use the architecture released by \citet{minigrid} to compute the observation embedding $e(o)$.
Specifically, this consists of a three 2D convolutional layers with ReLU activations and (input channels, output channels, kernel size) equal to (3, 16, (2, 2)), (16, 32, (2, 2)), and (32, 64, (2, 2)) respectively.
After the first convolutional layer is also a max pooling layer with kernel size (2, 2).
The output of the final convolutional layer is flattened and further passed through two fully connected layers with ReLU activations and output dimension 64.
We embed the $84 \times 84 \times 3$ map observation $f$ with an architecture derived from the standard DQN architecture from \citet{mnih2015human}, modified to take RGB images, rather than grayscale ones.
Specifically, this is computed as three convolutional layers with ReLU activations and (input channels, output channels, kernel size, stride) equal to (3, 32, (8, 8), 4), (32, 64, (4, 4), 2), and (64, 64, (3, 3), 1) respectively.
The output of the final convolutional layer is flattened and passed through a single fully connected layer of output dimension 64.

\textbf{Three-dimensional office variant.}
To compute embeddings $e(s)$ of states $s$ in the 3D office variant, we use the architecture released by \citet{gym_miniworld}.
Specifically, this consists of three convolutional layers with ReLU activations and (input channels, output channels, kernel size, stride) equal to (3, 32, (5, 5), 2), (32, 32, (5, 5), 2), and (32, 32, (4, 4), 2) respectively.
The output of these layers is flattened and passed through a single fully connected layer with output dimension 128.

\textbf{Probing experiment.}
In our probing experiment to test the agent's learned representation of the language floor plan, we begin by taking the agent's learned representation of an image of the language floor plan $e(s)$.
Then, we decode this learned representation into the tokens with a single LSTM layer with output dimension 100, followed by a final linear prediction layer with output dimension 64, equal to the vocab size.

\subsection{Hyperparameters}

\begin{table}[]
    \centering
    \begin{tabular}{cc}
        Hyperparameter & Value \\
        \toprule
        Discount Factor $\gamma$ & 0.99 \\
        Test-time $\epsilon$ & 0 \\
        Learning Rate & 0.0001 \\
        Replay buffer batch size & 32 \\
        Target parameters syncing frequency & 5000 updates \\
        Update frequency & 4 steps \\
        Grad norm clipping & 10
    \end{tabular}
    \caption{Hyperparameters shared across \dream, \rl, \import, and VariBAD.}
    \label{tab:hyperparameters}
\end{table}

We use the default hyperparameters for all methods reported in \citet{liu2021dream} without further tuning.
Specifically, the values used across all methods are given in Table~\ref{tab:hyperparameters}.
Additionally, all methods use lineary decaying $\epsilon$-greedy exploration, where the value of $\epsilon$ decays from 1 to 0.01 over 500000 steps for \rl, \import and VariBAD, and for \dream, the schedule is split into decaying from 1 to 0.01 in 250000 steps for both the exploration and task policies.

For the 2D office, we set the \dream penalty hyperparameter to $c = -0.1$, and use $c = 0$ in the 3D version, consistent with the MiniWorld experiments in \citet{liu2021dream}.
We set the encoder and decoder variances $\rho^2 = 0.1$ and information bottleneck weight $\lambda = 1$.

\end{document}